\documentclass[runningheads]{llncs}
  
\usepackage[export]{adjustbox}
\usepackage{hyperref}
\usepackage{booktabs}
\usepackage{graphicx}
\usepackage{subcaption}
\usepackage{cancel}
\usepackage{amsmath,amsfonts,amssymb,mathtools} 
\usepackage[misc,geometry]{ifsym}

\begin{document}

\title{Patch-based Brain Age Estimation from MR Images}

\author{Kyriaki-Margarita Bintsi \inst{1} \textsuperscript{(\Letter)}, Vasileios Baltatzis \inst{2}, Arinbj\"orn Kolbeinsson \inst{3}, Alexander Hammers \inst{2}, Daniel Rueckert \inst{1}}
\authorrunning{K. M. Bintsi et al.}

\institute{BioMedIA, Department of Computing, Imperial College London, UK \\
\email{m.bintsi19@imperial.ac.uk}\\ 
\and Biomedical Engineering and Imaging Sciences, Kings College London, UK \\
\and Department of Epidemiology and Biostatistics, Imperial College London, UK
}

\maketitle
\begin{abstract}
Brain age estimation from Magnetic Resonance Images (MRI) derives the difference between a subject's biological brain age and their chronological age. This is a potential biomarker for neurodegeneration, e.g. as part of Alzheimer's disease. Early detection of neurodegeneration manifesting as a higher brain age can potentially facilitate better medical care and planning for affected individuals. Many studies have been proposed for the prediction of chronological age from brain MRI using machine learning and specifically deep learning techniques. Contrary to most studies, which use the whole brain volume, in this study, we develop a new deep learning approach that uses 3D patches of the brain as well as convolutional neural networks (CNNs) to develop a localised brain age estimator. In this way, we can obtain a visualization of the regions that play the most important role for estimating brain age, leading to more anatomically driven and interpretable results, and thus confirming relevant literature which suggests that the ventricles and the hippocampus are the areas that are most informative. In addition, we leverage this knowledge in order to improve the overall performance on the task of age estimation by combining the results of different patches using an ensemble method, such as averaging or linear regression. The network is trained on the UK Biobank dataset and the method achieves state-of-the-art results with a Mean Absolute Error of 2.46 years for purely regional estimates, and 2.13 years for an ensemble of patches before bias correction, while 1.96 years after bias correction.

\end{abstract}

\section{Introduction}
Alzheimer's disease (AD) is a progressive neurodegenerative disease and the most common cause of dementia \cite{AlzheimersAssociation2019}. It has been demonstrated that AD results in atrophy of the brain\cite{savva2009age} as well as changes in the morphology of brain tissues. Specifically, atrophy firstly manifests itself in the hippocampus and it then expands through the cerebral cortex. Meanwhile, cerebrospinal fluid (CSF) spaces, i.e. the ventricles inside the brain and cisterns outside the brain, are enlarged. Biomarkers are an important diagnostic tool for detection of the early stages of AD \cite{davatzikos2011prediction}. A characteristic hallmark of neurodegenerative diseases such as AD is that they may result in accelerated aging of the human brain hence a promising biomarker that has been used in AD and other neuroimaging studies is brain age \cite{franke2019ten,cole2017predicting}, which measures the difference between the biological brain age of a subject and their real (chronological) age. 

Structural Magnetic Resonance (MR) images have been widely used for the measurement of brain changes related to age \cite{good2001voxel} and are often available in patients. For this reason, a considerable amount of work has been done on brain age estimation from MR images using machine learning (ML) techniques \cite{becker2018gaussian,franke2010estimating,tohka2016comparison}. The segmentation of the T1-weighted MR images into grey matter (GM), white matter (WM), and CSF regions, and region masks are usually used as input to a ML model \cite{kondo2015age,franke2010estimating}, achieving a Mean Absolute Error (MAE) of as low as 4.3 years \cite{liem2017predicting}. Recently, various CNN-based approaches which provide accurate brain age estimation using raw T1-weighted MR images with MAEs ranging from 3.3 to 4 years have been proposed \cite{jonsson2019brain,huang2017age,cole2017predicting}. In a recent competition on brain age estimation held last year (\href{https://web.archive.org/web/20200214101600/https://www.photon-ai.com/pac2019}{https://www.photon-ai.com/pac2019}) the top performing submission achieved results with a MAE of less than 3 years. In \cite{kolbeinsson2019robust}, the authors implemented a 3D ResNet model enhanced by a randomized tensor regression layer, achieving state-of-the-art results, and a MAE of 2.6 years. 

\begin{figure}[t]
\centering
\includegraphics[width = \textwidth]{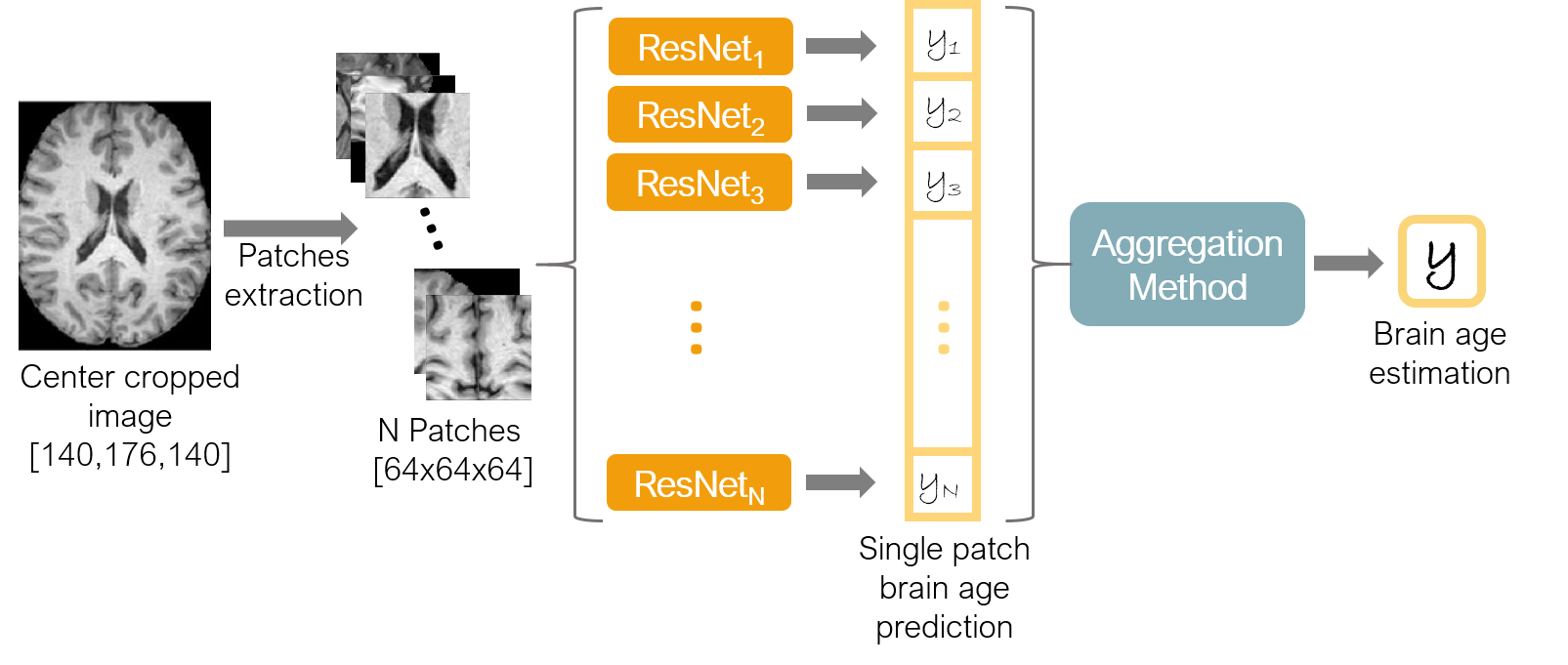}
\caption{Overview of the proposed method. $N$ 3D patches are being extracted from the center-cropped image. Every patch is used as input in a ResNet model that is trained to predict the subject's brain age, hence $N$ different age predictions $y_i$ are produced. The individual predictions are then combined using an ensemble method in order to achieve an improved estimation $y$.} \label{fig1}
\end{figure}

A common bias observed when estimating brain age, possibly because of regression dilution \cite{le2018nonlinear}, is the overestimation in younger subjects and the underestimation in older subjects. The term bias correction used in age estimation should not be confused with the bias field correction that is used as a pre-processing step for MRI images \cite{juntu2005bias}. There is a substantial amount of work \cite{smith2019estimation,cole2018brain,de2020commentary} going on towards the adjustment of the bias by applying a statistical bias correction method to the predicted ages. Recent works \cite{de2019population,beheshti2019bias} use the real age of the subject as a covariate to predict a bias-corrected age.

Patches have been widely used for brain disease prediction \cite{lian2018hierarchical,liu2018landmark} as they can provide information about subtle changes at a local level that are probably disregarded when working globally. To our knowledge, although patches could provide more localised estimations, they have not been used successfully for age regression. It is believed though that accurate localized predictions could provide interpretable insights \cite{pawlowski2019texture} that would be similar to clinical ones \cite{becker2018gaussian,cole2019multi}.

In this paper, we propose the use of patches of the brain, instead of the whole brain, as input and train independent 3D ResNet18 networks for every patch in order to obtain a regional estimate of brain age. Thus, we allow the network to focus on the specific patch without being affected by the rest of the brain. Furthermore, we evaluate which patches include important regions for the age regression and obtain a visualisation of these areas, thus making the results more interpretable and providing the first step for even more localised predictions. We, then, combine the results of the different patches to produce a more accurate brain age estimation for each subject, achieving state-of-the-art results with a MAE of 2.13 years before and 1.96 years after bias correction. 

\section{Materials \& Methods}

\subsection{Dataset and Pre-processing}
We use the UK Biobank (UKBB) dataset \cite{alfaro2018image} for the task of age regression. It incorporates a large collection of MRI scans of vital organs, including brain MR images, acquired with many different protocols such us T1-weighted and T2-weighted. The dataset used in this study contains 3D T1-weighted brain images, acquired from nearly 15,000 healthy subjects (ages: 44-73 years, females: 52.3\%). The images provided by the UKBB are sized 182x218x182 and are already skull-striped and non-linearly registered to MNI152 standard space, allowing anatomically consistent voxels among the subjects. We also normalize each brain scan to zero mean and unit variance.  From the 14,503 subjects provided by the UKBB, 753 did not provide their age and are removed from the study, thus leaving 13,750 3D brain images, 80\% of which are used for the training set, 15\% for the validation set and 5\% for the test set.

\subsection{Pipeline} 

An overview of the proposed method can be found in Figure \ref{fig1}. We first describe a baseline model which was used for the evaluation of our method. We then present the pipeline which led to state-of-the-art and more localized predictions.  

\textbf{Baseline Model} As a baseline for the evaluation of our method, we implement a version of a ResNet model \cite{he2016deep}, which uses 3D convolutions instead of 2D, similar to the one used in \cite{kolbeinsson2019robust}. 

\textbf{Patch-Based Prediction} In brief, we utilize patches of the brain, instead of the whole brain scans, with the purpose of understanding the effect of different parts of the brain on the predictions, and hence acquiring more localised predictions. We center crop each brain scan from 182x218x182 to 140x176x140. Afterwards, we extract 3D patches sized 64x64x64 with a stride of 38x56x38 from each brain, which leads to 27 patches per brain scan. We then train 27 18-layer 3D ResNet models, similar to the baseline one described before. We train, validate, and test on the same patch, thus on a specific region of the brain. This allows the network to focus on the specific patch independently of the rest of the image.

\begin{figure}[t]
\includegraphics[width = .5 \textwidth]{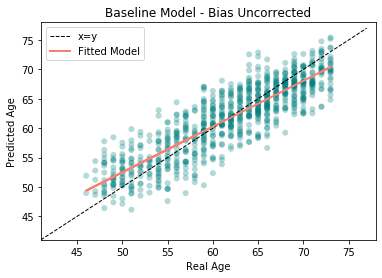}
\label{fig2a}
\includegraphics[width = .5 \textwidth]{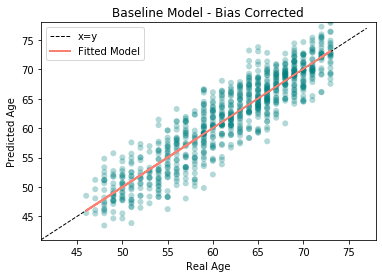}
\label{fig2b}\\
\caption{The predicted over the real age of the subjects for the baseline model before (a), and after bias correction (b). The black dashed line shows the line of identity ($x=y$), while the orange solid one shows the fitted model.} \label{fig2}
\end{figure}

\begin{table}[t]
\centering
\caption{Comparison of the performance of the most important experiments in terms of Mean Absolute Error (MAE) and $R^2$ score.}\label{tab1}
\begin{tabular}{|l|c|c|}
\hline
Experiment &  MAE (years) & $R^2$ score\\
\hline
Baseline model (uncorrected) &  2.64 & 0.77\\
Null model &  5.90 & 0.00\\
Best single-patch network &  2.46 & 0.80\\
Worst single-patch network & 4.19 & 0.14\\
Averaging (selected patches) & 2.26 & 0.84\\
\textbf{Linear Regression (all patches, uncorrected)} & \textbf{2.13} & \textbf{0.85}\\
\hline
Baseline model (corrected) & 2.43 & 0.80\\
\textbf{Linear Regression (all patches, corrected)} & \textbf{1.96} & \textbf{0.87}\\
\hline
\end{tabular}
\end{table}

\textbf{Ensemble model} We investigate whether the use of patches could lead to a more accurate brain age estimation. Therefore, we combine the 27 predictions of the single-patch networks with an ensemble model.

We experiment with two different ways for the fusion of the patch-based models. Firstly, we take the mean of the results of the multiple ResNets. So, if $N$ is the total number of patch-based models, $P \leq N$ is a subset of these models and $y_i$ is the prediction of every one of them, the prediction $y$ produced by the ensemble model is given by Equation \eqref{eq:1}: 
\begin{equation}
y = \frac{1}{P}  \sum\limits_{i=1}^P y_i
\label{eq:1}
\end{equation}

Moreover, since not all patches are equally important for age prediction, weighted averaging may produce better results. We, therefore, implement a linear regression model that predicts the brain age of each subject on the test set from $y_i$. The output prediction $y$ in this case is given by Equation \eqref{eq:2}:
\begin{equation}
y = w_0 +  \sum\limits_{i=1}^P w_i y_i
\label{eq:2}
\end{equation}
Here $w_0$ is the intercept term and $w_i$ are the learnable weights for each of the $P$ individual models, which are being learnt from the validation set. 

In addition, based on the previous experiments, we make some conclusions on which patches contain important features for brain age and which not. For this reason, we choose the number of the selected patch-based models $P$ so that each individual $MAE < threshold$ on the validation set. We choose that threshold heuristically. We investigate both averaging the predictions and using a meta-regressor similar to the one described above.  

\textbf{Bias Correction} The bias correction technique adopted in this work is the one proposed by \cite{beheshti2019bias}. We fit the relationship of the brain age delta and the chronological age using Equation \eqref{eq:3}: 
\begin{equation}
\Delta = \alpha * Y_{chronological} + \beta
\label{eq:3}
\end{equation}
where $Y_{chronological}$ is the real age of the subject and $\Delta$ the brain age delta function. The $\alpha$ and $\beta$ represent the slope and intercept respectively and are then used for the estimation of the corrected predicted age from Equation \eqref{eq:4}: 
\begin{equation}
Y_{corrected} = Y_{predicted} - (\alpha * Y_{chronological} + \beta)
\label{eq:4}
\end{equation}

\textbf{Training} We train the network using backpropagation \cite{rumelhart1985learning}, with batch size 8 for 40 epochs. We use a mean squared error loss for the age regression. Furthermore, we use the adaptive moment estimation (Adam) optimizer \cite{kingma2014adam} with a learning rate lr=0.0001. 
The experiments are implemented on an NVIDIA Titan RTX using Pytorch deep-learning library \cite{paszke2019pytorch}. 

\section{Results}

Three different experiments are implemented. For the evaluation of the age regression task in every model, MAE and  the $R^2$ coefficient are computed. The main results of the experiments can be found in Table \ref{tab1}. 

\textbf{Baseline Model} As depicted in Figure \ref{fig2}, the network that uses the whole images as an input achieves a MAE of 2.64 years and an $R^2$ score of 0.77. Comparing to \cite{kolbeinsson2019robust}, our model performs better to their simple 3D ResNet, perhaps because we used a later release of the UKBB that includes around double the number of subjects. When bias correction is applied, the performance improves, with a MAE of 2.43 years and an $R^2$ score of 0.80 observed. 

In addition, we create a model that predicts the average age of the population for every subject with the purpose of the comparison of the results of this null model with our results. The model achieves a MAE of 5.90 years. 

\begin{figure}[t]
\begin{subfigure}{0.5\textwidth}
\includegraphics[scale = .6, center]{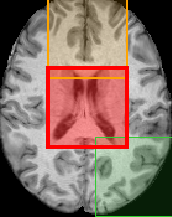}
\label{fig3a}
\end{subfigure}
\begin{subfigure}{0.5\textwidth}
\includegraphics[width = 0.8\linewidth]{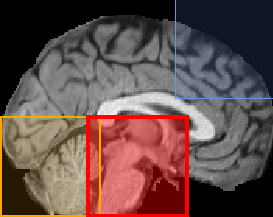}
\label{fig3b}\\
\end{subfigure}
\caption{Transverse (a) and mid-sagittal (b) slices of a brain MR image with three colored patches for each case. The colors of the patches indicate the performance of this single-patch model in terms of Mean Absolute Error (MAE). Specifically: red patches indicate a MAE of less than 3 years, orange patches between 3 and 3.5 years, green patches between 3.5 and 4 years, and blue patches greater than 4 years. The patches that perform best include the ventricles (a), and the hippocampus (b) (hippocampus further lateral than the plane shown).} \label{fig3}
\end{figure}

\textbf{Patch-based Predictions} When we use a single patch as input, the results are highly dependent on the area of the brain where the patch is extracted from. We train 27 ResNets for 27 patches, achieving a MAE from around 2.5 to 4.2 years. The patches with the best performance are the ones that include the ventricles or the hippocampus, with MAEs of less than 2.5 and 2.7 years, respectively, and $R^2$ score of 0.80 and 0.76, respectively, as well as patches that include parts of these regions. On the contrary, patches that contain substantial amounts of background perform much worse with a MAE of 3.5-4.2 years and an $R^2$ score of 0.15-0.6. In Figure \ref{fig3}, various patches, colored based on the performance of the corresponding models, are illustrated as part of a whole brain slice.

\textbf{Ensemble Model} By combining the single-patch models, we achieve an improved performance. Concerning the aggregation of the individual models, apart from the average of all the patches, we also calculate the average of the 6 patches that achieve greater performance on the validation set ($MAE < 3$ years), hence the ones that include important information on brain aging. The MAE is 2.66 and 2.26 years, respectively, and the $R^2$ score 0.78 and 0.83, respectively.

\begin{figure}[t]
\includegraphics[width = .5 \textwidth]{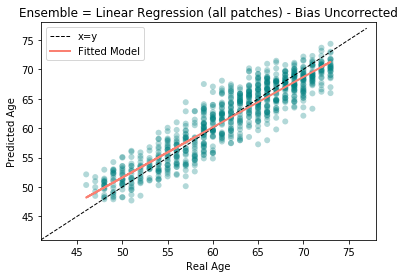}
\label{fig4a}
\includegraphics[width = .5 \textwidth]{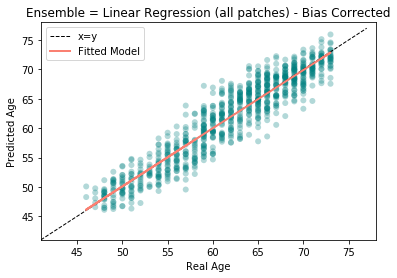}
\label{fig4b}\\
\caption{The predicted over the real age of the subjects when the predicted ages result from a linear regression of all the patches before bias correction (a), and after bias correction (b). The black dashed line shows the line of identity ($x=y$), while the orange solid one shows the fitted model.} \label{fig4}
\end{figure}

The use of linear regression works better. Specifically, we achieve a MAE of 2.13 years when all the single-patch models are used. Furthermore, when we use only the models of the patches with the most important information for the age of the brain,  we achieve a MAE of 2.16 years. Regarding $R^2$ score, it is 0.85 in both cases. In this case, bias correction is applied as well and we achieve a MAE of 1.96 years, and an $R^2$ score of 0.87. The scatter plots of the real over predicted ages, before and after bias correction is applied, are shown in Figure \ref{fig4}.

\section{Discussion}
A plethora of deep learning methods based on CNNs which use whole brain images as input has been used for estimating brain age via regression, achieving good performance (MAE=2.6 years) \cite{kolbeinsson2019robust}. In this study, we have explored how patches of the brain can be used for this task, by training a network for each patch that tries to find robust but patch-specific features for the age estimation. Moreover, we combine the results of the different networks via an ensemble approach with the purpose of a more accurate estimation. The results of the main experiments can be found in Table \ref{tab1}.

Based on the results, we conclude that the patches that include parts of the ventricles and the hippocampus tend to predict the age of the subject more accurately with a MAE of less than 3 years, thus our results are consistent with the existing literature \cite{savva2009age}. The MAE in some of these cases is lower than that for the baseline network which utilizes the volume of the whole brain, proving that not all parts of the brain are needed for an accurate prediction. Therefore, less computational effort is required for comparable results. On the other hand, patches that include part of the background do not perform as accurately. We observe that although the brain shrinks while age increases, the networks do not manage to capture the changes in the cerebral cortex as accurately. An explanation for this could be the way the images are registered, which reduces size differences between brains, or the relatively small range of existing ages in the dataset, especially of ages beyond 65 years when atrophy tends to accelerate \cite{coupe2017towards}. Furthermore, the use of patches provides a visualization of the parts of the brain that are valuable for age prediction, leading to a more understandable prediction. Although the patches are big and do not provide detailed localization, this work is the first step towards an interpretable and localized brain age estimation. As our next step, we will use smaller patches (e.g. sized 32x32x32 or 16x16x16), either from the whole brain or from the best-performing patches. More localised detection of a deviation from normal trajectories may also be useful for the automatic detection of more focal pathologies such as ventricular enlargement in hydrocephalus or hippocampal atrophy in temporal lobe epilepsies \cite{keihaninejad2012classification}.

We also improve the age prediction by combining the results of a few weak networks. Firstly, we do so by simply averaging the results of all the independent networks. However, this does not lead to any improvement on the predictions (MAE = 2.66 years). Thus, we then leverage the knowledge acquired from the previous experiment and average only the predictions from patches that seemed to include the most relevant information about brain age. There is a major boost in the regression performance (MAE = 2.26 years), showing that by combining results of weaker networks we can achieve a better prediction. Moreover, we aggregate the single-patch models by using a linear regression, as we believe that by giving appropriate weights to the results of every patch, better results could be achieved. Interestingly, in this case, using all the individual models performs similarly (MAE = 2.13 years) to when we use the selected ones (MAE = 2.16 years). The reason is, probably, that by learning the weights, the linear regression model can suppress the effect of the least accurately performing models, while promoting the best performing ones. This results in an age estimator which is more accurate than the baseline model as well as the existing literature. In future work, we intend to integrate the ensemble method with the training of the sub-networks in order to accommodate an end-to-end training. 

After bias correction, we notice a considerable increase of the performance of the network (MAE = 1.96 years). Although using the chronological age as a covariate for the bias correction is not meaningful for a ML model, it could prove important clinically for giving insights for the subject's brain age, as demographics, including chronological age, are usually known. 

\section{Conclusion}

In this work, we propose the use of patches of the brain, instead of the whole 3D brain scan, as input to separate 3D ResNets for brain age estimation. We combine the results of the networks with an ensemble method, such as linear regression. The method provides accurate results outperforming the state-of-the-art. In addition, we provide localized predictions by specifying the patches that perform better on the task. Precise age prediction in healthy participants, e.g. from UKBB as in this study, can now be used as a baseline to compare trajectories in different pathologies such as Alzheimer's \cite{coupe2017towards}. 

\subsubsection{Acknowledgements} KMB would like to acknowledge funding from the EPSRC Centre for Doctoral Training in Medical Imaging (EP/L015226/1).

\bibliographystyle{splncs04}
\bibliography{paper14}

\end{document}